\documentclass[11pt, a4paper,tikz,x11names]{article}
\usepackage{lrec}
\usepackage{graphicx}
\usepackage{tabularx}
\usepackage{soul}

\usepackage{epstopdf}
\usepackage[latin1]{inputenc}

\usepackage{hyperref}
\usepackage{xstring}
\usepackage[table,xcdraw]{xcolor}
\usepackage{multirow}
\usepackage[edges]{forest}

\title{Multi-class Multilingual Classification of Wikipedia Articles\\Using Extended Named Entity Tag Set}
\name{Hassan S. Shavarani\textsuperscript{*}\thanks{*
The author was an intern at AIP Center for Advanced Intelligence, during this project.} \\
  School of Computing Science \\
  Simon Fraser University, BC, Canada \\
  {\tt sshavara@sfu.ca} \\\and
  \large{\textbf{Satoshi Sekine}}\\
  AIP Center for Advanced Intelligence \\
  Riken, Tokyo, Japan \\
  {\tt satoshi.sekine@riken.jp} \\
}

\address{}

\abstract{
Wikipedia is a great source of general world knowledge which can guide NLP models better understand their motivation to make predictions. Structuring Wikipedia is the initial step towards this goal which can facilitate fine-grain classification of articles.
In this work, we introduce the \textit{Shinra 5-Language Categorization Dataset} (\textsc{SHINRA-5LDS}), a large multi-lingual and multi-labeled set of annotated Wikipedia articles in Japanese, English, French, German, and Farsi using \textit{Extended Named Entity} (ENE) tag set. We evaluate the dataset using the best models provided for ENE label set classification and show that the currently available classification models struggle with large datasets using fine-grained tag sets.}

\begin{document}
\begin{NoHyper}

\maketitleabstract
\end{NoHyper}

\section{Introduction}

Major progress has been made in different tasks in Natural Language Processing, yet our models are still not able to describe why they make their decisions when summarizing an article, translating a sentence, or answering a question. Lack of meta information (e.g. general world knowledge regarding the task) is one important obstacle in the construction of language understanding models capable of reasoning about their considerations when making decisions (predictions).

Wikipedia is a great resource of world knowledge for human beings, but lacks the proper structure to be useful for the models. To address this issue and make a more structured knowledge-base, Sekine et al. \shortcite{sekine2018shinra} try to structure Wikipedia. Their final goal is to have, for each Wikipedia article, known entities and sets of attributes, with each attribute linking to other entities wherever possible. The initial step towards this goal would be to classify the entities into predefined categories and verify the results using human annotators\footnote{
Please note that the verification process plays an important role in the knowledge-base construction process since it leads to what is represented to our models as world facts.}.

Throughout the past years, many have tried classifying Wikipedia articles into different category sets mostly containing between 3 to 15 class types \cite{toral2006proposal,watanabe2007graph,dakka2008augmenting,chang2009wikisense,tardif2009improved}. Such categorization type sets are not much helpful when the classified articles are being used as the training data for question answering systems, since the extracted knowledge-base does not provide detailed enough information to the model.

On the other hand, much larger categorization type sets such as Cyc-Taxonomy \cite{lenat1995cyc}, Yago-Taxonomy \cite{yago}, or Wikipedia's own taxonomy of categories \cite{schonhofen2009identifying} are not suitable for classifying Wikipedia articles since the tags are not verifiable for annotators\footnote{They need to keep 200K+ classes in mind to find the most suitable ones for the article at hand or verify the classifier category prediction for it.}. In addition, taxonomies are not designed in a tree format, so some categories might have multiple super-categories and this would make the verification process much harder for articles discussing multiple topics.

Considering the mentioned problem requirements, we believe \textit{Extended Named Entities Hierarchy} \cite{sekine2002}, containing 200 fine-grained categories tailored for Wikipedia articles, is the best fitting tag set.

\begin{table*}[ht]
\setlength\tabcolsep{2.5pt}
\begin{center}

\begin{tabular}{c||c|c|c|c||c||c|c||c}\label{table:dataset}
\multirow{2}{*}{language} & \multicolumn{4}{c||}{average size in folds} & \multirow{2}{*}{\begin{tabular}[c]{@{}c@{}}total\\classes\end{tabular}} & \multicolumn{2}{c||}{average count}     & \multirow{2}{*}{\begin{tabular}[c]{@{}c@{}}max \\ annotations\end{tabular}} \\ \cline{2-5} \cline{7-8}
& train & dev & test & total & & article/class & annot./article &\\ \hline\hline
\rowcolor[HTML]{E0E0E0}
ja  & 96,321.8   & 12,004.9   & 12,006.3  & 120,333  & 141 & 853.426 & 1.0359 & 5\\
en & 42,652.8   & 5,301.1    & 5,301.1   & 53,228   & 127 & 419.331 & 1.0359 & 5\\ 
\rowcolor[HTML]{E0E0E0}
fr  & 27,750.5   & 3,425.7    & 3,424.8   & 34,601   & 113 & 306.204 & 1.0347 & 5\\
de  & 23,969.8   & 2,958.8    & 2,959.4   & 29,888 & 108 & 276.741 & 1.0309 & 5\\ 
\rowcolor[HTML]{E0E0E0}
fa  & 11,329.4   & 1,388  & 1,386.6   & 14,104   & 80  & 176.3 & 1.0342& 5\\
\end{tabular}
\caption{Statistics about \textit{Shinra 5-Language Categorization Dataset} as well as the suggested average train/dev/test size of the data sectors used in the benchmark experiments.}
\end{center}
\vspace{-2em}
\end{table*}

Higashinaka et al. \shortcite{higashinaka2012} were the first to use this extended tag set as output labels when categorizing Wikipedia pages. Their model was trained using a hand-extracted feature set that converted the pages into model compatible input vectors. Following their work, Suzuki et al. \shortcite{suzuki2016} augmented the extracted input features with trained vectors modelling the links between different Wikipedia pages. They proposed a more complex model for learning the mapping between the converted articles and the labels. Although providing useful insights, neither have considered exploring the multi-lingual nature of many Wikipedia articles. 

We base this work on Sekine et al. \shortcite{shinra}'s work in which they have hired linguists as annotators and educated them on the Extended Named Entities (ENE) tag set to annotate each article with up to 6 different ENE classes. We exploit the Wikipedia language links in the annotated articles to create our multi-lingual Wikipedia classification dataset. Section 2 details our dataset creation process.

We then use the models suggested by Higashinaka et al. \shortcite{higashinaka2012} and Suzuki et al. \shortcite{suzuki2016}, the only works close enough to our task at hand (to the best of our knowledge), to benchmark the created dataset. Section 3 provides more details about our multi-lingual feature selection method and the models. Section 4 presents our experimental setup and the classification results.

%
\section{Dataset Collection and Annotation}


Recently, Sekine et al. \shortcite{shinra} created an annotated dataset containing 782,517 Japanese Wikipedia articles in different areas, covering 175 out of 200 ENE labels\footnote{The rest of the categories were not covered since they did not find any articles under the category which could meet the selection criteria at the time.}. The articles are selected from Japanese Wikipedia with the condition of being hyperlinked at least 5 times from other articles in Wikipedia. They had instructed annotators\footnote{Majority of the annotators were post secondary degree holders in linguistics.} to label the collected articles using at most 6 labels\footnote{They report no inter annotator agreement data, but report that 200 samples from the data have been randomly selected and passed to skilled annotators to validate/verify the quality of the annotations.} from the 200 suggested ENE labels\footnote{The data is provided and maintained for \textsc{SHINRA2020-ML} classification task. The latest version of it is available via \href{http://shinra-project.info/shinra2020ml/}{http://shinra-project.info/shinra2020ml/}}.

We considered a subset of the annotated articles which have been hyperlinked at least 100 times (as Suzuki et al. \shortcite{suzuki2016} suggest) that led to a 120,333 Japanese Wikipedia articles (annotated with 141 out of 200 ENE labels and maximum 5 annotations per article).
We collected the content of the same article titles in English, French, German, and Farsi Wikipedia sections\footnote{
The wikidump data used for extracting the articles' content was the May 20, 2018 snapshot of Wikipedia in all five languages. 
}, relying only on Wikipedia language links. Language links connect the articles representing the exact same topic from one language to another. We used the labels assigned to Japanese version of the articles to all the articles in other four languages (in case any existed), since ENEs are language agnostic and the pages offered the same content.

To perform the language link exploration, we first created the graph of language links for all the (``wikipedia id'', ``language'') pairs linking one article in one of the five languages to another article in another language. We also took into account the Wikipedia redirect links in our exploration process, since sometimes language links connect articles to redirect pages in other languages.
Using the language links graph, we formed ``Entities'' grouping all different (``wikipedia id'', ''language'') pairs representing the same subject and then applied the ENE labels to the articles in different languages.

We call this multi-lingual multi-labeled collection of Wikipedia articles, the ``\textit{Shinra 5-Language Categorization Dataset}'' (\textsc{SHINRA-5LDS})\footnote{The data (available at \href{http://shinra-project.info/download/}{shinra-project.info/download/}) will only contain the (``wikipedia id'', ``language'') pairs and can be combined with the actual articles (in wiki-dumps) using wikipedia\_id references.}, and we release the dataset alongside this paper to enable the other researchers to perform the benchmark on multi-labeled Japanese, English, French, German, and Farsi Wikipedia categorization using their suggested methods. 
Table \ref{table:dataset} contains the total number of annotated articles in each of the languages as well as the total number of ENE classes with at lease one article annotated in that class,
the average number of articles collected in each class, and the average number of annotations assigned to each article by the annotators. 
%
%
%
%
%
%
%
%
%
%
%
%
%
\section{Feature Selection and Models} \label{sec:models}
To perform the benchmark, we surveyed the available suggested models for multi-class categorization of Wikipedia articles and selected the models suggested by Higashinaka et al. \shortcite{higashinaka2012} and Suzuki et al. \shortcite{suzuki2016}, since both have suggested classifying Wikipedia articles using ENEs. 
We also decided to study the usefulness of the hierarchy in the process of training the classifiers using ENEs. Hence, we also selected the models suggested by Wehrmann et al. \shortcite{wehrmann2018} as our third set of models. The following sections describe our feature selection procedure and briefly explain each of the models.

\subsection{Feature Selection}

A fair comparison between the models on the dataset is not possible unless we can guarantee the same input to each of them. With that in mind, we went through the feature selection methods suggested in \cite{wang2012}, \cite{higashinaka2012} and \cite{suzuki2016} and created a union of their suggestions. 

However, we had to remove some of the features such as `\textit{Last one/two/three characters in the headings or titles}" or ``\textit{Last character type (Hiragana/Katakana/Kanji/Other)}" from the union due to the multi-lingual nature of our task.

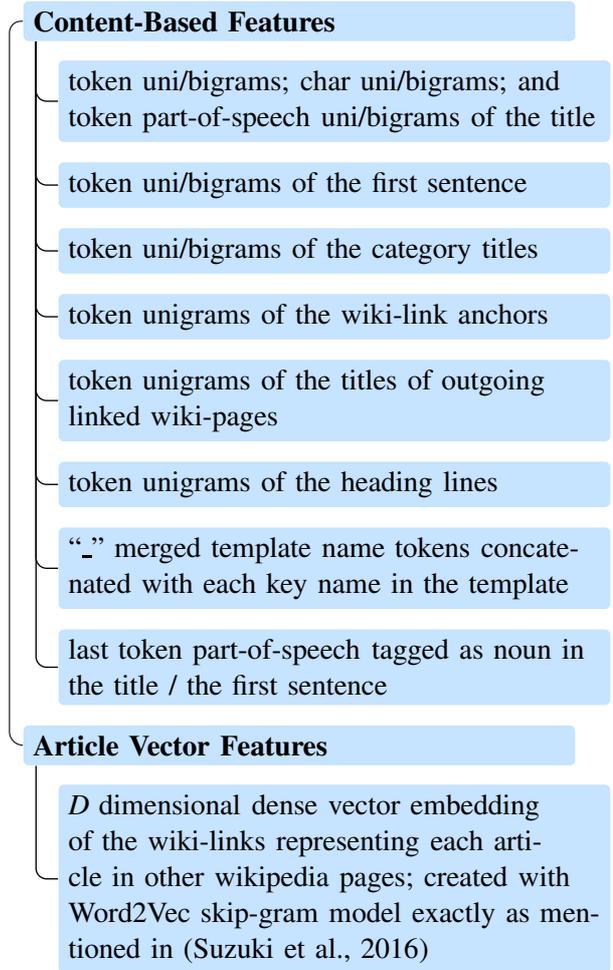
\begin{figure}[ht]
    \begin{forest}
      my label/.style={
        label={right:{#1}},
      },
      for tree={
        folder,
        text=black,
        minimum height=0.50cm,
        text width=70mm,
        if level=0{fill=white}{fill/.wrap pgfmath arg={SlateGray1}{int(4-(mod((level()-1),4)))}},
        rounded corners=2pt,
        grow'=0,
        edge={rounded corners,line width=0.5pt},
        fit=band,
      },
      [,phantom
        [\bf Content-Based Features , name=feat1
          [token uni/bigrams; char uni/bigrams; and token part-of-speech uni/bigrams of the title]
          [token uni/bigrams of the first sentence]
          [token uni/bigrams of the category titles]
          [token unigrams of the wiki-link anchors]
          [token unigrams of the titles of outgoing linked wiki-pages]
          [token unigrams of the heading lines]
          [``\_'' merged template name tokens concatenated with each key name in the template]
          [last token part-of-speech tagged as noun in the title / the first sentence]
        ]
        [\bf Article Vector Features, name=feat2
          [\textit{D} dimensional dense vector embedding of the wiki-links representing each article in other wikipedia pages; created with Word2Vec skip-gram model exactly as mentioned in \cite{suzuki2016}]
        ] {
        \draw[-,rounded corners] (feat2.west)--++(-5pt, 1pt)--++(0pt, 272pt)--(feat1.west);
        }
      ]
    \end{forest}
    \caption{Features extracted from each article}
    \label{fig:featureset}
\end{figure}

Figure \ref{fig:featureset} summarizes the final unified schema for categorization of the Wikipedia articles in \textsc{SHINRA-5LDS}.

\subsection{Binary Logistic Regression}

Higashinaka et al. \shortcite{higashinaka2012} suggested learning a set of separate \textit{Binary Logistic Regression Classifier Models} to learn the contribution of the extracted features towards the final selected class. 
We employ this model to indicate the classification difficulty level of our dataset using a simple model.

\subsection{Joint-NN and Joint-NN++}
Suzuki et al. \shortcite{suzuki2016} suggested that combining all the separate Logistic Regression Classifier Models into a \textit{2-Layer Perceptron Neural Network} may result in capturing more information for better confidence in assigning ENE classes to the articles. They call their suggested model \textit{Joint-NN} and conclude that their model is better in learning the correlation of the extracted features with the output ENE labels than a separate set of logistic regression models or even a separate set of 2-Layer Perceptron Networks each of which trying to predict one of the labels.
We employ their suggested \textit{Joint-NN} model and also try augmenting it with another additional layer (we call the augmented model \textit{Joint-NN++}) in our benchmark experiments.

\subsection{Hierarchical Multi-Label Classification Networks}
To examine the extent of information lying in the Hierarchy of ENEs, we propose using \textit{Hierarchical Multi-Label Classification Networks (HMCN)}.
Wehrmann et al. \shortcite{wehrmann2018} suggest two different settings for the HMCNs both of which perform the prediction of the label hierarchy in a top-down manner. 
The first setting, \textit{HMCN Feed-forward (HMCN-F)}, uses a separate explicit part of the network for predicting each level of the hierarchy. On the other hand, \textit{HMCN Recurrent (HMCN-R)} learns of the hierarchy by recurrently feeding the prediction of the previous top layer to the next lower level in hierarchy.
We suggest to employ \textit{HMCN-R} in addition to \textit{HMCN-F} to examine the effect of model compression on learning to predict the hierarchy of ENEs at test time.

\begin{table*}
\begin{tabular}{l||l|l|l|l|l||l|l|l|l|l}
\multicolumn{1}{c||}{\multirow{2}{*}{Model}} & \multicolumn{5}{c||}{dev}                                                                                                       & \multicolumn{5}{c}{test}                                                                                                      \\ \cline{2-11} 
\multicolumn{1}{c||}{}                       & \multicolumn{1}{c|}{ja} & \multicolumn{1}{c|}{en} & \multicolumn{1}{c|}{de} & \multicolumn{1}{c|}{fr} & \multicolumn{1}{c||}{fa} & \multicolumn{1}{c|}{ja} & \multicolumn{1}{c|}{en} & \multicolumn{1}{c|}{de} & \multicolumn{1}{c|}{fr} & \multicolumn{1}{c}{fa} \\ \hline\hline
\rowcolor[HTML]{E0E0E0}
{} & {} & {} & {} & {} & {} & {} & {} & {} & {} & {}\\
\rowcolor[HTML]{E0E0E0}
\multirow{-2}{*}{\begin{tabular}[l]{@{}l@{}}Binary Logistic\\Regression\end{tabular}} & \multirow{-2}{*}{\begin{tabular}[c]{@{}c@{}}71.25\end{tabular}} & \multirow{-2}{*}{\begin{tabular}[c]{@{}c@{}}76.24\end{tabular}} & \multirow{-2}{*}{\begin{tabular}[c]{@{}c@{}}69.56\end{tabular}} & \multirow{-2}{*}{\begin{tabular}[c]{@{}c@{}}69.74\end{tabular}} & \multirow{-2}{*}{\begin{tabular}[c]{@{}c@{}}79.70\end{tabular}} & \multirow{-2}{*}{\begin{tabular}[c]{@{}c@{}}71.18\end{tabular}} & \multirow{-2}{*}{\begin{tabular}[c]{@{}c@{}}72.69\end{tabular}} & \multirow{-2}{*}{\begin{tabular}[c]{@{}c@{}}69.27\end{tabular}} & \multirow{-2}{*}{\begin{tabular}[c]{@{}c@{}}65.83\end{tabular}} & \multirow{-2}{*}{\begin{tabular}[c]{@{}c@{}}66.45\end{tabular}} \\

Joint-NN \textsuperscript{\dag} & \textbf{80.19} & 78.43 & \textbf{81.58} & 81.23 & 79.71 & 77.31 & 78.18 & \textbf{81.41} & 78.85 & 76.34\\
\rowcolor[HTML]{E0E0E0}
Joint-NN++ & 77.73 & \textbf{81.13} & 79.88 & \textbf{83.53} & \textbf{85.25} & \textbf{77.40} & \textbf{80.80} & 79.88 & \textbf{83.43} & \textbf{79.78}\\
HMCNF & 72.07 & 73.59 & 71.43 & 73.54 & 76.07 & 71.25 & 73.31 & 69.71 & 70.22 & 75.83\\
\rowcolor[HTML]{E0E0E0}
HMCNR & 61.63 & 64.28 & 64.66 & 64.80 & 70.45 & 61.38 & 63.04 & 61.70 & 64.65 & 70.20\\
\end{tabular}\caption{The classification accuracy of the predicted labels. Partially correct labels have also contributed partially to the scores.\newline
\textsuperscript{\dag} 
Despite our endeavor to keep the settings comparable to the original model, comparison between our results and theirs would not be fair, since the size of datasets used in our experiments and also the number of classes are different than theirs.}\label{table:results}
\end{table*}

\subsection{Training and Evaluation}
To preform the multi-label classification, we suggest passing all the model predicted membership distributions through a Sigmoid layer and assign the label to the article if the predicted probability after passing through Sigmoid is above 0.5.

The evaluation measure would then be the micro-averaged precision \cite{sorower2010} of the predicted labels. In addition, to prevent the domination of more frequent classes on the training procedure, we suggest weighted gradient back-propagation. The back-propagation weight of each article would be calculated using $w = \frac{N}{\sum_{n=1}^{N}{f(l_n)}}$ where $N$ is the number of labels assigned to the article (with a maximum of 6) and $f(l_n)$ counts the total train-set articles to which label $l_n$ has been assigned.
The loss function used for training all the models has been \textit{Binary Cross Entropy Loss} averaged over all the possible classes.

\section{Experiments and Results}
We implemented all the models suggested in \S\ref{sec:models} using the PyTorch framework\footnote{\href{https://pytorch.org}{https://pytorch.org - v0.4.1}}. For part-of-speech tagging the title and first sentences of the articles mentioned in the feature selection schema (Figure \ref{fig:featureset}) and also normalization and tokenization of the articles, we used Hazm Toolkit\footnote{\href{https://github.com/sobhe/hazm}{https://github.com/sobhe/hazm}} for Farsi, Mecab Toolkit \cite{kudo2006} for Japanese, and TreeTagger Toolkit\footnote{\href{https://github.com/miotto/treetagger-python}{https://github.com/miotto/treetagger-python}} for English, French, and German. 

In all of our experiments, we have used Adam optimizer \cite{kingma2015} with a learning rate of $1e^{-3}$ and have performed gradient clipping \cite{pascanu2013} of 5.0. We have initialized all of the network parameters with random values between $(-0.1, 0.1)$.
We have done training on mini-batches of size 32, and to have a fair comparison, all the experiments have been conducted with 30,000 steps (batches) of randomly shuffled training instances to train the model parameters. The hidden layer size of all the models in each layer has also been set to 384\footnote{We have also tried larger sizes of hidden layers for simpler models but the results did not vary much, so we removed the probability of difference in learning capability of the models in different parameter set sizes from our experiment result analysis.}.

We have performed the evaluation in a 10 fold cross validation manner in each fold of which 80\% of the data has been used for training, 10\% for validation and model selection, and 10\% for testing. In addition, classes with a frequency less than 20 in the dataset have been ignored in the train/test procedure.

Table \ref{table:results} depicts the benchmarked micro-averaged precision of classification prediction of the articles in our dataset.
The results initially demonstrate that the dataset is not an easy one as the Binary Logistic Regression model is not achieving very high accuracy scores. Besides, the lower scores for Japanese in comparison to the other languages demonstrate the higher difficulty level of classification for a larger category set size for all the models.

On the other hand, the consistency of the superior results of non-hierarchical models to the hierarchical models shows that the leaf-node ENEs contain all the necessary information to perform the classification over them, and the hierarchy may only add more confusion to model decisions.

Last but not least, the overall precision scores depict that the currently available models struggle with larger more complex annotated sets of Wikipedia articles.

In our future studies, we will focus on providing more complex models which can capture more information from the articles (leading to better classification scores) and we will also focus on using the results of our classifier to create a bigger structured knowledge-base to augment the currently available NLP models.

\section{Bibliographical References}
\label{main:ref}

\bibliographystyle{lrec}
\bibliography{refs}


\end{document}